# *Self-Replicating Machines in Continuous Space with Virtual Physics*

Arnold Smith[*], Peter Turney[†] (corresponding author), Robert Ewaschuk[‡]

## Abstract

JohnnyVon is an implementation of self-replicating machines in continuous two-dimensional space. Two types of particles drift about in a virtual liquid. The particles are automata with discrete internal states but continuous external relationships. Their internal states are governed by finite state machines but their external relationships are governed by a simulated physics that includes brownian motion, viscosity, and spring-like attractive and repulsive forces. The particles can be assembled into patterns that can encode arbitrary strings of bits. We demonstrate that, if an arbitrary "seed" pattern is put in a "soup" of separate individual particles, the pattern will replicate by assembling the individual particles into copies of itself. We also show that, given sufficient time, a soup of separate individual particles will eventually spontaneously form self-replicating patterns. We discuss the implications of JohnnyVon for research in nanotechnology, theoretical biology, and artificial life.

[*] Institute for Information Technology, National Research Council of Canada, Ottawa, Ontario, Canada, K1A 0R6, arnold.smith@nrc.ca, (613) 991-1147

[†] Institute for Information Technology, National Research Council of Canada, Ottawa, Ontario, Canada, K1A 0R6, peter.turney@nrc.ca, (613) 993-8564 (corresponding author)

[‡] School of Computer Science, University of Waterloo, Waterloo, Ontario, Canada, N2L 3G1, raewasch@uwaterloo.ca

# 1  Introduction

John von Neumann is well known for his work on self-replicating cellular automata [19]. His ultimate goal, however, was to design self-replicating physical machines, and cellular automata were simply the first step towards the goal. Before his untimely death, he had sketched some of the other steps. One step was to move away from the discrete space of cellular automata to the continuous space of classical physics (pages 91-99 of [19]). Following the path sketched by von Neumann, we have developed JohnnyVon, an implementation of self-replicating automata in continuous two-dimensional space.

JohnnyVon consists of a virtual "soup" of two types of particles that drift about in a simulated liquid. The particles are automata with discrete internal states that are regulated by finite state machines. Although the particles are internally discrete, the external relationships among the particles are continuous. Force fields mediate the interactions among the particles, enabling them to form and break bonds with one another. A pattern encoding an arbitrary string of bits can be assembled from these two types of particles, by bonding them into a chain. When a soup of separate individual particles is "seeded" with an assembled pattern, the pattern will replicate itself by assembling the separate particles into a new chain.[1]

The design of JohnnyVon was inspired by DNA and RNA. The individual automata are intended to be like codons and the assembled patterns are like strands of DNA or RNA. The simulated physics in JohnnyVon corresponds (very roughly) to the physics inside cells. The design was also influenced by our interest in nanotechnology. The automata in JohnnyVon can be seen as tiny nanobots, floating in a liquid vat, assembling structures in a manufacturing plant. Another source of guidance in our design was, of course, the research on self-replicating cellular automata, which has thrived and matured greatly since von Neumann's pioneering work. In particular, although the broad outline of JohnnyVon is derived from physics and biology, the detailed design of the system borrows much from automata theory. The basic entities in JohnnyVon are essentially finite automata, although they move in a continuous space and are affected by smoothly-varying force fields.

---

[1] The copies are mirror images; however, that is not a problem. This point is discussed in Section 5.

We discuss related work in Section 2. JohnnyVon is closely related to research on self-replicating cellular automata. A significant difference is that the automata in JohnnyVon are mobile. There is some related work on mobile automata that move in a two-dimensional space, including Tim Hutton's Squirm3 [6], which has many similarities to JohnnyVon. The work of Lionel and Roger Penrose, who created self-replicating machines using pieces of plywood, is also relevant [11].

JohnnyVon is described in detail in Section 3. The motion of the automata is determined by a simulated physics that includes brownian motion, viscosity, and spring-like attractive and repulsive forces. The behaviours and interactions of the automata are determined by a small number of internal states and state transition rules.

We present two experiments in Section 4. First, we show that an arbitrary seed structure can assemble copies of itself, using individual automata as building blocks. Second, we show that a soup of separate individual automata can spontaneously form self-replicating structures, although we have deliberately designed JohnnyVon so that this is a relatively rare event.

Section 5 is our interpretation of the experiments. Section 6 is concerned with limitations of JohnnyVon and future work. Some potential applications of this line of research are given in Section 7 and we conclude in Section 8.

## 2  Related Work

JohnnyVon is related to research in self-replicating automata, mobile automata, and physical models of self-replication. Self-replicating machines are also related to self-assembling machines and universal constructors.

### 2.1    Self-Replicating Cellular Automata

Since von Neumann's pioneering work [19], after a hiatus, research in self-replicating cellular automata is now flourishing [7], [13], [15], [17], [18]. Most of this work has involved two-dimensional cellular automata. A two-dimensional grid of cells forms a discrete space, which is infinite and unbounded in the abstract, but is necessarily finite in a computer implementation. The cells are (usually identical) finite state machines, in which a cell's state depends on the states of its neighbours, according to a set of (deterministic) state transition rules. The system begins with each cell in an initial state (chosen by the user) and the states change synchronously in discrete time steps. With carefully selected transition rules, it is possible to create self-replicating patterns. The initial states of the cells are "seeded" with a certain pattern (usually the pattern is a small loop). Over time, the pattern spreads from the seed to nearby cells, eventually filling the available space.

Although this work has influenced and guided us, JohnnyVon is different in several ways. The automata in JohnnyVon are (essentially) finite state machines, but they are mobile, rather than being locked in a grid. The automata move in a continuous two-dimensional space, rather than a discrete space (but time in JohnnyVon is still discrete). The states of the automata are mainly discrete and finite, but each automaton has a position and a velocity, and the force fields around the tips of each particle have smooth gradients, all of which are represented with floating point numbers.[2] The movements of the automata are governed by a simple simulated physics. We claim that these differences from self-replicating cellular automata make JohnnyVon more realistic, and thus more suitable for guiding work on self-replicating nanotechnology and research on the origins of life and related issues in theoretical biology. Aside from the increased realism, JohnnyVon is interesting simply because it is significantly different from cellular automata.

## 2.2    Mobile Automata

There has been other research on mobile automata (e.g., bio-machines [8] and generalized mobile automata [23]), combining Turing machines with cellular automata. Turing machines move from cell to cell in an *N*-dimensional grid space, changing the states of the cells and possibly interacting with each other. However, as far as we know, the only investigation of self-replicating mobile automata, other than JohnnyVon, is Hutton's Squirm3 [6].

The Squirm3 simulation uses artificial chemistry to support self-replicating molecules. Virtual atoms occupy cells in a two-dimensional discrete grid space. The atoms move randomly in the grid space, presumably due to virtual brownian motion, like the particles in JohnnyVon. When two atoms occupy adjacent cells, they may form a bond with each other, again like the particles in JohnnyVon. If a seed molecule, consisting of a string of bonded atoms, is placed in a soup of free atoms, it will replicate itself by a series of virtual chemical reactions with the free atoms.

Whereas Squirm3 uses virtual chemistry, JohnnyVon uses virtual physics; whereas Squirm3 uses a discrete grid space, JohnnyVon uses a continuous space; whereas the elementary units in Squirm3 are virtual atoms, the elementary units in JohnnyVon are virtual machines (e.g., nanobots). Otherwise, the two systems have many similarities, although they were developed independently.[3]

---

[2] We discuss in Section 3.3 the extent to which the automata in JohnnyVon are finite state machines.

The choice of a chemical model versus a physical model depends on the task at hand. Perhaps a chemical model is more suitable for investigating the origins of life, but a physical model is more suitable for designing self-replicating nanobots. However, some models of the origin of life, such as Cairn-Smith's clay model [1], might be better approached with a model like JohnnyVon, rather than a model like Squirm3.

### 2.3    Physical Models of Self-Replication

Lionel Penrose, with the help of his son, Roger, made actual physical models of self-replicating machines, using pieces of plywood [11]. His models are similar to JohnnyVon in several ways. Both involve simple units that can be assembled into self-replicating patterns. In both, the units move in a continuous space. Another shared element is the harnessing of random motion for replication. JohnnyVon uses simulated brownian motion and Penrose required the plywood units to be placed in a box that was then shaken back and forth. Penrose described both one-dimensional and two-dimensional models, in which motion is restricted to a line or to a plane.

An obvious difference between the Penrose models and JohnnyVon is that the former are physical whereas the latter is computational. One advantage of a computational model is that experiments are perfectly repeatable, given the same initial conditions and the same random number seed. Another advantage is the ability to rapidly modify the computational model, to explore alternative models.

---

[3] We only discovered each other's work after our respective systems were implemented.

A limitation of the Penrose models is that the basic units are all identical, so they cannot use binary encoding. They could encode information by length (the number of units in an assembled pattern), but the mechanism for ensuring that length is replicated faithfully is built in to the physical structure of the units. Thus altering the length involves building new units. On the other hand, JohnnyVon can encode an arbitrary binary string without making any changes to the basic units.

### 2.4  Self-Assembly versus Self-Replication

Self-assembly is the coordinated action of independent entities under local (distributed) control to produce a larger structure [21], [22]. Typically, components spontaneously self-assemble by moving about randomly in a liquid or gas until a stable minimum-energy configuration is achieved. Self-assembly occurs in living organisms (in growth and development) and has also been demonstrated in chemistry (with self-assembling molecular structures) and in computational simulations. Programmable molecular self-assembly has been demonstrated experimentally. For example, research in DNA computation has led to the development of synthetic DNA double-crossover molecules that self-assemble into two-dimensional crystals [20]. Research in nanotechnology often involves self-assembly, but self-assembly is not limited to molecular-scale structures; in principle, self-assembly can occur at any scale.

Self-replication is the process whereby an object or structure makes a copy of itself. Sipper distinguishes self-replication from self-reproduction [17]. In self-replication, an exact duplicate is made. In self-reproduction, genetic operators such as crossover and mutation result in offspring that are different from their parents.

Self-assembly does not necessarily imply self-replication, since the larger structure that is produced by self-assembly may be quite different from the original structure. Likewise, self-replication does not necessarily imply self-assembly, since self-replication might use a centralized (non-local) control system. JohnnyVon, however, involves both self-assembly and self-replication.

## 2.5    Universal Construction versus Self-Replication

Von Neumann sketched five distinct models for self-replication: the kinematic model, the cellular automata model, the excitation-threshold-fatigue model, the continuous model, and the probabilistic model [19]. Only the cellular automata model was described in detail. All five models had a common architecture, with two parts: a universal computer and a universal constructor. Together, these two parts could construct anything that could possibly be constructed (with cellular automata), given the appropriate set of instructions. Thus self-replication in von Neumann's models arose as a special case of universal construction. Von Neumann's approach to self-replication is an example of self-replication without self-assembly, since control is centralized in the universal computer. JohnnyVon is related to von Neumann's kinematic and continuous models, but JohnnyVon is not a universal constructor, since it is only capable of self-replication.

# 3   JohnnyVon

We begin the description of JohnnyVon with an informal discussion of the model (3.1). We then define some terminology (3.2), followed with an outline of the states of the automata (3.3). The next subsection considers the attractive and repulsive fields that govern the interactions among the automata (3.4) and the following subsection sketches the simulated physics (3.5). The sixth subsection explains how the automata decide when to split apart (3.6) and the seventh subsection discusses the treatment of time in JohnnyVon (3.7). The final subsection is concerned with the implementation of JohnnyVon (3.8).

## 3.1   Informal Discussion

Much (but not all) of the previous work on self-replication has been based on two-dimensional cellular automata [17]. This research has produced some very interesting results, but the practical implications are not clear. For example, Sayama's structurally dissolvable self-replicating loop is interesting for its elegant and simple design, but it is quite different from self-replication in biology [15], [16]. It seems that many features of its design were based on the constraints of cellular automata. We believe that the assumption of a discrete space is the aspect of cellular automata models that is least realistic and most problematic for extrapolating from results with self-replicating automata to applications in biology and nanotechnology. This belief was a central motivation for the design of JohnnyVon.

A continuous space computer simulation is generally much more computationally intensive than a comparable discrete space simulation. When we began our work, it was not obvious to us that today's computers would be sufficiently powerful for a continuous space simulation of self-replicating machines. Our goal was to demonstrate self-replication and self-assembly in continuous space with virtual physics, using a desktop computer to run the simulation. The constraints on our design were that the basic components should be physically plausible as miniature machines and that the control system should be purely local. Cellular automata models of self-replication are local, but they lack physical plausibility. In the real world, self-assembly and self-replication invariably require moving chunks of matter around in a continuous space, and cellular automata are not a convenient way to model this.

For example, in Hutton's system, atoms move around in a discrete space and they can form self-replicating chains, but he found it necessary to force an atom to stop moving in certain situations [6]. An atom cannot move if it is bonded to another atom and moving would take it out of the neighborhood of the atom to which it is bonded. This results in groups of atoms (i.e., molecules) that become frozen in the grid space and cannot move until some of their bonds are broken. This is not physically plausible, and it is representative of the kinds of constraints that are imposed by the assumption of discrete space.

The design of JohnnyVon was based on the idea that strings (chains) of particles, of arbitrary length, should be capable of forming spontaneously, and once formed, they should be relatively stable. Each particle is a T-shaped structure. Particles form strings by bonding together at the tips of the horizontal arms of the T structures. Strings replicate by attracting randomly floating particles to the tips of the vertical arms of the T structures and holding them in place until they join together to form a replica of the original string.

Bonds between particles in a string can be broken apart by brownian motion (due to random collisions with virtual molecules of the containing fluid) and by jostling from other particles. Particles can also bond to form a string by chance, without a seed string, if two T structures meet under suitable conditions. Strings that are randomly broken or formed can be viewed as mutations. We intentionally designed JohnnyVon so that mutations are relatively rare, although they are possible. Faithful replication is intended to be much more common than mutation.

The attractive fields around particles have limited ranges, which can shrink or expand. This is one of the mechanisms that we use to ensure faithful replication. The fields shrink when we want to discourage bonding that could cause mutations and the fields expand when we want to encourage bonding that should lead to faithful replication.

A mechanism is needed to recognize when a string has attracted and assembled a full copy of itself. Without this, each seed string would attract a single copy, but the seed and its copy would remain bonded together forever. Therefore the automata send signals to their neighbours in the string, to determine whether a full copy has been assembled. When the right signal is received, a particle releases the bond on the vertical arm of the T structure and pushes its corresponding particle away.

To create a simulation that will run on a desktop computer, we have made some design decisions that limit the physical plausibility of the model. For example, our attractive and repulsive fields were inspired by electrical fields, but they do not strictly obey the physics of electromagnetism. This is discussed in more detail in Section 6.

### 3.2    Definitions

The following definitions will facilitate our subsequent discussion. To better understand these definitions, it may be helpful to look ahead to Table 1 and Figure 1.

**Codon:** a T-shaped object that can encode one bit of information. There are two types of codons, type 0 codons and type 1 codons.

**Container:** the space that contains the codons. Codons move about in a two-dimensional continuous space, bounded by a grey box. The centers of the codons are confined to the interior of the grey box.

**Liquid:** a virtual liquid that fills the container. The trajectory of a codon is determined by brownian motion (random drift due to the liquid) and by interaction with other codons and the walls of the container. The liquid has a viscosity that dampens the momentum of the codons.

**Soup:** liquid with codons in it.

**Field:** an attractive or repulsive area associated with a codon. The range of a field is indicated by a coloured circle. In addition to attracting or repelling, a field can also exert a straightening force, which twists the codons to align their arms linearly. The fields behave somewhat like springs.

**Arm:** a black line in a codon that begins in the middle of the codon and ends in the center of a coloured field.

**Tip:** the outer end of an arm, where the red, blue, green, and purple fields are centered.

**Middle:** the inner ends of the arms, where the three arms join together. This is not the geometrical center of the codon, but it is treated as the center of mass in the physical simulation.

**Red (blue, purple, green) arm:** an arm that ends in a red (blue, purple, green) field.

**Bond:** two codons can bond together when the field of one codon intersects the field of another. Not all fields can bond. This is described in detail later.

**Red (blue, purple, green) neighbour:** the codon that is bonded to the red (blue, purple, green) arm of a given codon.

**Single strand:** a chain of codons that are red and blue neighbours of each other.

**Double strand:** two single strands that are purple and green neighbours of each other.

**Small (large) field:** a field may be in one of two possible states, small or large. These terms refer to the radius of the circle that delimits the range of the field.

**Free codon:** a codon with no bonds.

**Time:** the number of steps that have been executed since the initialization of JohnnyVon. The initial configuration is called step 0 (or time 0).

### 3.3  States

The state of a codon in JohnnyVon is represented by a vector, rather than a scalar. The state vector of a codon has 15 elements:

- 3 floating point variables for position (x-axis location, y-axis location, angle)
- 3 floating point variables for velocity (x-axis velocity, y-axis velocity, angular velocity)

- 4 binary variables for field size (four fields per codon, two states per field)

- 3 whole-number-valued variables (one for each arm of the codon) for "pointers" that identify the codon to which a given arm is bonded (if any)

- 1 three-valued variable for *strand_location_state*

- 1 three-valued variable for *splitting_state*

These elements are described in more detail in the following subsections.

The six-dimensional finite-valued sub-vector, consisting of the four binary variables and the two three-valued variables, is the part of each codon that corresponds to the traditional notion of a finite state machine. This six-dimensional sub-vector has a total of $2^4 \times 3^2 = 144$ possible states.[4]

Chou and Reggia demonstrated the emergence of self-replicating structures in two-dimensional cellular automata, using 256 states per cell [2]. The state values were divided into "data fields", which were treated separately. In other words, Chou and Reggia used state vectors, like JohnnyVon, rather than state scalars. We agree with Chou and Reggia that state vectors facilitate the design of state machines.

The remaining nine variables in the state vector (position, velocity, bond pointers) represent information about external relationships among codons, rather than the internal states of the codons. For example, the absolute position of a codon is not important; the interactions of the codons are determined by their relative positions. These nine variables are analogous to the grid in cellular automata. As far as internal states alone are concerned, the codons are finite state machines. It is their external relationships that make the codons significantly different from cellular automata.

---

[4] Some combinations of states are not actually possible. See Tables 2 to 5.

Sayama notes that his nine-state, five-neighbour cellular automata model is probably the simplest cellular automata model for simulating evolution of self-replicators that has been proposed so far, with a rule space of size $9^5 = 59,049 \approx 6 \times 10^4$ [16]. Since each codon in JohnnyVon can have at most three neighbours (one codon bonded to each of the three arms; four neighbours including the given codon itself), the rule space is of size $144^4 = 429,981,696 \approx 4 \times 10^8$. This is more complex than Sayama's model but considerably less complex than Chou and Reggia's 256-state, nine-neighbour cellular automata model, with a rule space of size $256^9 = 2^{72} \approx 5 \times 10^{21}$ [2]. On the other hand, JohnnyVon's continuous space with virtual physics is much more complex than the discrete grid space of cellular automata. In a sense, with JohnnyVon, we have shifted some of the complexity out of the internal structure of each automaton and into the external structure of relationships among the automata.

### 3.4  Fields

The codons have attractive and repulsive fields, as shown in Table 1. These fields determine how codons interact with each other to form strands. There are five types of fields, which we have named according to the colours that we use to display the codons in JohnnyVon's user interface (purple, green, blue, red, yellow). All five fields have two possible states, called *large* and *small*, according to the radius of the circle that delimits the range of the field. All fields in a free codon are small. Note that the codons are asymmetric: every codon has its red arm on the left side and its blue arm on the right side (when viewed as in Table 1, with the purple or green arm pointing up). This never changes during a run, since there is no way for codons to rotate out of the 2D plane.

Insert Table 1 here.

Table 2 gives the state transition rules for the field states. Fields switch between small and large as bonds are formed and broken.

> Insert Table 2 here.

The interactions among the fields are listed in Table 3. Fields can pull codons together, push them apart, or twist them to align their arms.

> Insert Table 3 here.

### 3.5　Physics

JohnnyVon runs in a sequence of discrete time steps. Each codon has a position (x-axis location, y-axis location, and angle) and a velocity (x-axis velocity, y-axis velocity, and angular velocity) in two-dimensional space. Although time is measured in whole numbers, position and velocity are represented with floating point numbers. (We discuss time in more detail in Section 3.7.) Each codon has one unit of mass.

The internal state changes of a codon are triggered by the presence and state of neighbouring codons. One codon "senses" another when it comes within range of one of its force fields. It could be said that the force fields are also sensing fields, and when one of these fields expands, its sensing ability expands equally.

We think of the container as holding a thin layer of liquid, so although the space is two-dimensional, codons are allowed to slide over one another. This simplifies computation, since we do not need to be concerned with detecting collisions between codons. It also facilitates replication, since free codons can move anywhere in the container, so it is not possible for them to get trapped behind a wall of strands.

It is interesting to note that strands are emergent structures that depend only on the local interactions of individual codons. There is no data structure that represents strands; each codon is treated separately and interacts only with its immediate neighbours.

### 3.5.1   Brownian Motion

Codons move in a virtual liquid. Brownian motion is simulated by applying a random change to each codon's (linear and angular) velocity at each time step. This random velocity change may be thought of as the result of a collision with a molecule of the liquid, but we do not explicitly model the liquid's molecules in JohnnyVon.

### 3.5.2   Viscosity

We implement a simple model of viscosity in the virtual liquid. With each time step, a codon's velocity is adjusted towards zero by multiplying the velocity by a fractional factor. One factor is applied for x-axis and y-axis velocity (linear viscosity) and another factor is applied for angular velocity (angular viscosity).

### 3.5.3   Attractive Force

When two codons are bonded, an attractive force pulls their bonded arms together. This force acts like a spring joining the tips of the bonded arms. The strength of the spring force increases linearly with the distance between the tips, until the distance is greater than the sum of the radii of the fields, at which point the bond is broken. The spring force modifies both the linear and angular velocities of the bonded codons. (The angular velocity is modified because the force acts on the codon tip, rather than on the center of mass.)

### 3.5.4   Repulsive Force

The repulsive force also acts like a spring, joining the centers of the yellow fields, pushing the codons apart. The strength of the spring force decreases linearly with the distance between the centers of the yellow fields, until the distance is greater than the sum of the radii of the fields, at which point the force ceases. The spring force modifies both the linear and angular velocities of the bonded codons.

### 3.5.5   Straightening Force

When two codons are bonded, a straightening force twists them to align their bonded arms linearly. This force is purely rotational; it has no linear component. The two bonded codons rotate about their middles, so that their bonded arms lie along the line that joins their middles. The straightening force for a given codon is linearly proportional to the angle between the bonded arm of the given codon and the line joining the middle of the given codon to the middle of the other codon.

### 3.5.6   Spring Dampening

The motion due to the attractive and straightening forces is dampened, in a way that is similar to the viscosity that is applied to brownian motion. The dampening prevents unlimited oscillation. No dampening is applied to the repulsive force, since oscillation is not a problem with repulsion. The linear velocities of a bonded pair of codons are dampened towards the average of their linear velocities, by a fractional factor (linear dampening). The angular velocities of a bonded pair of codons are dampened towards zero, by another fractional factor (angular dampening).

## 3.6   Splitting

When a complete double strand forms, the yellow fields switch to their large states and split the double strand into two single strands. The decision to split is controlled by a purely local process. Each codon has an internal state that is determined by the states of its neighbours. When a codon enters a certain internal state, its yellow field switches to the large state. The splitting is determined by a combination of two state variables, the *strand_location_state* and the *splitting_state*.

The *strand_location_state* has three possible values:

> 0 =   Initial state: I do not think I am at the end of a (possibly incomplete) double strand.

    1 =    I think I might be located at the end of a (possibly incomplete) double strand.

    2 =    My green or purple neighbour also thinks it might be at the end of a (possibly incomplete) double strand.

An incomplete double strand occurs when a single strand has partially replicated. On one side, there is the original single strand, and, attached to it, there are one or more single codons or shorter single strands. The state transition rules for *strand_location_state* are designed so that a codon can only be in state 2 when it is at one of the two extreme ends of a (complete or incomplete) double strand.

The *splitting_state* also has three possible values:

    x =    Initial state: I am not ready to split.

    y =    I am ready to split.

    z =    I am now splitting.

A codon's yellow field switches to the large yellow state when *splitting_state* becomes z.

The following tables give the rules for state transitions. Table 4 lists the rules for *strand_location_state* and Table 5 provides the rules for *splitting_state*.

Insert Table 4 here.

Insert Table 5 here.

### 3.7 Time

For the discrete elements in the state vector, there is a natural relation between changes in state and increments of time, when each unit of time is one step in the execution of JohnnyVon. For the continuous elements in the state vector (position and velocity), the time scale is somewhat arbitrary. The physical rules that are used to update the position and velocity are continuous functions of continuous time. In a computational simulation of a continuous process, it is necessary to sample the process at a succession of discrete intervals. In JohnnyVon, the parameter *timestep_duration* determines the temporal resolution of the simulation. The parameter may be seen as determining how finely continuous time is sliced into discrete intervals, or, equivalently, it may be seen as determining how much action takes place from one step of the simulation to the next. Changing the value of the parameter is equivalent to rescaling the magnitudes of the forces.

A small value for *timestep_duration* yields a fine temporal resolution (i.e., a small amount of action between steps) and a large value yields a coarse temporal resolution. If the value is too small, the simulation will be computationally inefficient; many CPU cycles will be wasted on making the simulation unnecessarily precise. On the other hand, if the value is too large, the simulation may become unstable; the behaviour of the objects in the simulation may be a very poor approximation to the intended continuous physical dynamics.

The actual value that we use for *timestep_duration* has no meaning outside of the context of the (arbitrary) values that we chose for the magnitudes of the various physical parameters (force field strengths, viscosity, brownian motion, etc.; see the appendix for details). We set *timestep_duration* by adjusting it until it seemed that we had found a good balance between computational efficiency and physical accuracy.

When comparing different runs of JohnnyVon, with different values for *timestep_duration,* we found it useful to normalize time by multiplying the number of steps by *timestep_duration.* For example, if you halve the value of *timestep_duration,* then half as much action takes place from one time step to the next, so it takes twice as many time steps for a certain amount of action to occur. Therefore, as JohnnyVon runs, it reports both the number of time steps and the normalized time (the product of the step number and *timestep_duration*). However, in the following experiments, we only report the number of steps, since the normalized time has no meaning when taken out of context.

### 3.8    Implementation

JohnnyVon is implemented in Java. The source code is available under the GNU General Public License (GPL) at http://purl.org/net/johnnyvon/.

We originally implemented JohnnyVon in C++. The current version is in Java because we found it difficult to make the C++ version portable across different operating systems. Informal testing suggests that the Java version runs at about 75% of the speed of the C++ version. We believe that the slight loss of speed in the Java version is easily offset by the gain in portability and maintainability.

## 4    Experiments

In our first experiment, we seed a soup of free codons with a pattern and show that the pattern is replicated. In the second experiment, we show that a soup of free codons, given sufficient time, will spontaneously generate self-replicating patterns.

### 4.1    Seeded Replication

In Figure 1, Images 1 to 7 show a typical run of JohnnyVon with a seed strand of eight codons and a soup of 80 free codons. (Image 8 in Figure 1 is for Section 4.2.) Over many runs, with different random number seeds, we have observed that the seed strand reliably replicates.

Insert Figure 1 here.

**Image 1, Step 250:** This image shows JohnnyVon near the start of a run, after 250 steps have passed. A soup of free codons (randomly located) has been seeded with a single strand of eight codons (placed near the center). The strand of eight codons encodes the binary string "00011001" (0 is purple, 1 is green). In the strand, the red fields overlap with the corresponding blue fields of the red neighbours. To show the overlapping fields, the half of the circle that is closest to the codon with the red arm is coloured red, and the half of the circle that is closest to the codon with the blue arm is coloured blue.

**Image 2, Step 6,325:** Six codons have bonded with the seed strand, but they have not yet formed any red-blue bonds.

**Image 3, Step 18,400:** Eight codons have bonded with the seed strand, but these eight codons have not yet formed a complete strand. One red-blue bond is missing. These bonds can only form when the red and blue arms meet linearly to within ± π/256 radians. This happens very rarely when the codons are drifting freely, but it happens dependably when the codons are held in position by the purple-green bonds.

**Image 4, Step 22,000:** The eight codons formed their red-blue bonds, making a complete strand of eight codons. This caused the yellow fields in the double strand to switch to their large states, breaking the bonds between the two single strands and pushing them apart. In this image, the yellow fields are still large. After a few more time units have passed, they will return to their small states. Note that the seed strand encodes "00011001", but the daughter strand encodes "01100111". This is discussed in Section 5.

**Image 5, Step 25,950:** We now have two single strands, and they have started to form bonds with the free codons.

**Image 6, Step 30,850:** The daughter strand has replicated itself, producing a granddaughter. The original seed strand and the granddaughter encode the same bit string.

**Image 7, Step 127,950:** There are only a few free codons left. Eventually they will bond with the strands, leaving a stable soup of partially completed double strands. The elapsed real-time from step 250 (Image 1) to step 127,950 (Image 7) was approximately 45 minutes on an Intel Pentium III running at 600 MHz.

### 4.2   Spontaneous Replication

JohnnyVon was intentionally designed so that self-replicating patterns can arise from free codons without a seed, but only rarely. It is difficult for red-blue bonds to form, due to the narrow angle at which the arms must meet (± π/256 radians – see Table 2). These bonds are very unlikely to form unless the codons are held in position by green-purple bonds. However, given sufficient time, two free codons will eventually come into contact in such a way that a red-blue bond is formed and a self-replicating strand of length two is created. Image 8 in Figure 1 shows an example.

**Image 8, Step 164,450:** A strand of length two has spontaneously formed from a soup of 88 free codons. Very shortly after forming, it replicated. The elapsed real-time from step 0 to step 164,450 was about 15 minutes. This is less real-time per step than the previous experiment because there are fewer calculations when there are no bonds between the codons.

## 5  Interpretation of Experiments

The first experiment shows that a pattern containing arbitrary information can replicate itself. Note that all codon interactions in JohnnyVon are local; no global control system is needed. (This is also true of the various implementations of self-replicating cellular automata.)

It is apparent in Image 7 that (approximately) half of the single strands are mirror images of the original seed strand (in Image 1). More precisely, let *X* be an arbitrary sequence of 0s and 1s that we want to encode. Let $n(X)$ be the string that results when every 0 in $X$ is replaced with 1 and every 1 in $X$ is replaced with 0. Let $r(X)$ be the string that results when the order of the characters in $X$ is reversed. When a strand with the pattern $X$ replicates, the resulting new strand will have the pattern $r(n(X))$. Therefore, if we seed a soup of free codons with a pattern $X$, then the final result will consist of about 50% strands with the pattern $X$ and 50% with the pattern $r(n(X))$.

Penrose anticipated this problem [11]. He suggested it could be avoided by making the pattern symmetrical. Let $c(X, Y)$ be the string that results when the string $Y$ is concatenated to the end of string $X$. Let $g(X)$ be $c(X, r(n(X)))$. Note that $g(X)$ is equal to its negative mirror image, $r(n(g(X)))$. That is, if $g(X)$ replicates, the resulting string is exactly $g(X)$ itself. The function $g(X)$ enables us to encode any arbitrary string $X$ in such a way that replication will not alter the pattern. 100% of the final strands will be copies of $g(X)$.

The second experiment shows that self-replicating patterns can spontaneously arise. The strands in this case are of length two, but it is possible in principle for mutations to extend the length of the strands (although we have not observed this).

Strands of length two have an evolutionary advantage over longer strands, since they can replicate faster. On rare occasions, when running JohnnyVon with a seed of length eight (as in Section 4.1), a strand of length two has spontaneously appeared. The length-two strand quickly out-replicates the length-eight strand and soon predominates.

We have intentionally designed JohnnyVon so that its most likely behaviour is to faithfully replicate a given seed strand. However, we have allowed a small possibility of red-blue bonds forming without a seed pattern, which allows both spontaneous generation of self-replication and mutation of existing strands. (The probability of mutation can be increased or decreased by adjusting the red-blue bonding angle above or below its current value of ± π/256 radians.) Since there is selection (for rapid replication), JohnnyVon supports a limited degree of evolution: there is inheritance, mutation, and selection.[5]

Cellular automata can also support self-replication [6], [7], [13], [17], [19], evolution [2], [6], [13], [14], [16], [19] and spontaneous generation of self-replication without seeding [2], [6]. The novelty in JohnnyVon is that these three features appear in a computer simulation that includes continuous space and virtual physics. We believe that this is an important step towards building physical machines with these features.

One reason that von Neumann turned to cellular automata models was that the computers of his time were not sufficiently powerful to simulate his kinematic or continuous models [19].[6] JohnnyVon demonstrates that computer hardware has advanced to a level where we can now simulate other models of self-replication, besides cellular automata models. Perhaps one of the more important lessons we have learned from this work is that simulation of self-replication in continuous space with virtual physics is now feasible. When we began our work, it seemed quite possible that we could fail to achieve our goals, due to hardware limitations.

## 6  Limitations and Future Work

One area we intend to look at is the degree to which the internal codon states can be simplified while still exhibiting the basic features of stability and self-replication. We make no claim that we have found the simplest codon structure that will exhibit the intended behaviours.

---

[5] Selection is due to the different replication rates of the competing strands. Selection can only continue indefinitely if there is an endless source of free codons and there is room for unlimited population growth. In a finite container, selection will eventually halt, unless a mechanism is added to JohnnyVon, to supply free codons and remove excess strands.

JohnnyVon contains only genotypes (genetic code) with no phenotypes (bodies). The only evolutionary selection that JohnnyVon currently supports is selection for shorter strands, since they can replicate faster than longer strands. In order to support more interesting selection, we would like to introduce phenotypes. In natural life, DNA can be read in two different ways. During reproduction, DNA is copied verbatim, but during growth, DNA is read as instructions for building proteins.[7] We would like to introduce this same distinction into a future version of JohnnyVon. One approach would be to add new "protein" particles to complement the existing codon particles. Free protein particles would bond to a strand of codons, which would act as a template for assembling the proteins. Once a string of proteins has been assembled, it would separate from the codon strand and then it would fold into a shape, analogous to the way that real proteins fold. To achieve interesting evolution, the environment could be structured so that certain protein shapes have an evolutionary advantage, which somehow results in increased replication for the corresponding codon strands. Previous work on simulating cells may be useful for this project [3], [5].

Another limitation of JohnnyVon is the simplistic virtual physics. In many cases, we sacrificed some physical plausibility in the design of JohnnyVon in order to achieve the goals of computational tractability and self-replication. For example, electrostatic attraction and repulsion in the real world have an infinite range, but all of the fields in JohnnyVon have quite limited ranges (relative to the size of the container). Our codons can only interact when their fields are in contact with one another, so it is not necessary to calculate the forces between every pair of codons. This significantly reduces the computation, especially when there are many free codons, since the trajectory of a free codon is determined solely by brownian motion and viscosity.

---

[6] However, even his cellular automata model of self-replication was only recently implemented [12].

[7] This is a bit of an oversimplification, which ignores issues such as sexual reproduction, where the copying is not verbatim.

In addition to the limited range of the attractive and repulsive forces, there are the different types (colours) of fields, the rules for interactions among fields, the ability to switch between small and large states for fields, and the spring-like behaviours of the fields. These aspects of the design of JohnnyVon represent trade-offs that we made to achieve our goals; they do not represent properties of self-replication in the real world.

However, it is likely possible to significantly increase the physical realism of JohnnyVon without sacrificing speed. This is another area for future work. It may be that the direction taken will depend on the application. The changes that would make JohnnyVon more realistic for a biologist, for example, may be different from the changes that would be appropriate for a nanotechnologist.

Finally, it may be worthwhile to develop a 3D version of JohnnyVon. The current 2D space might be insufficiently realistic for some applications.

## 7  Applications

JohnnyVon was designed with nanotechnology in mind. We hope that it may some day be possible to implement the codons in JohnnyVon (or some distant descendant of JohnnyVon) as nanomachines. We imagine that the two types of codons could be mass produced by some kind of macroscopic manufacturing process, and then sprinkled in to a vat of liquid. A seed strand could be dropped in the vat, and the nanomachines would quickly replicate the seed. This imaginary scenario might never become reality, but the success of the experiments in Section 4 lends some plausibility to this project.

Merkle has argued that a central objective of nanotechnology is to make products inexpensively, and that self-replication is an effective approach to very low cost manufacturing [9], [10]. This view is shared by several others in the field [4]. Nanotechnologists who are doing research in self-replicating machines will certainly want to simulate the machines before they actually build them. We believe that such simulations, to be realistic, will necessarily include continuous space and virtual physics.

JohnnyVon may also contribute to theoretical biology, by increasing our understanding of natural life. As Penrose mentioned, models of this kind may help us to understand the origins of life on Earth [11]. Simulations like JohnnyVon, using continuous space and virtual physics, seem well-suited to further elaboration and testing of theories such as Cairn-Smith's clay model of the origin of life [1].

## 8  Conclusion

JohnnyVon includes the following features:

- automata that move in a continuous 2D space

- self-replication of seed patterns

- spontaneous generation of self-replication from free codons (self-replication without seed patterns)

- evolution (inheritance, mutation, and selection)[8]

- virtual physics (brownian motion, viscosity, attraction, repulsion, dampening)

- the ability to encode arbitrary bit strings in self-replicating patterns

- local interactions (no global control structures)

JohnnyVon is the first computational simulation to combine all of these features.

Von Neumann sketched a path that begins with self-replicating cellular automata and ends with self-replicating physical machines. We agree with von Neumann that, at some point along this path, it is necessary to move away from discrete space models, towards continuous space models. JohnnyVon is such a step.

---

[8] Evolution is limited, as discussed in Section 5.

## Acknowledgements

Thanks to the anonymous referees of *Artificial Life* for their very helpful comments. Thanks to Tim Hutton for sharing his draft paper with us.

## Appendix

To facilitate the replication of our experimental results, Table 6 shows the internal parameters of JohnnyVon and the values that were used in the experiments. The source code is available, as discussed in Section 3.8.

Insert Table 6 here.

## List of Tables

## List of Figures

Table 1. The two types of codons and their field states. The fields of a free codon are always small. The fields become large only when codons bond together, as described in the next table. A codon's fields may be in a mixture of states (one field may be small and another large). Note that the circles are not drawn to scale, since the small fields would be invisibly small at this scale.

|  | Type 0 codons (purple codons) | Type 1 codons (green codons) |
|---|---|---|
| All fields in their small states | | |
| All fields in their large states | | |

Table 2. State transitions in fields. Fields can change from small to large or vice versa, but they never change colour.

| Current field state | Next field state | Transition rules |
|---|---|---|
| small red | large red | If a small blue field touches a small red field and the arms of their respective codons are aligned linearly to within ± $\pi/256$ radians, then both fields switch to their large states and their codons are designated as being bonded together. As long as the two fields continue to intersect, at any angle, they remain in the bonded state, and any third field that intersects with the two fields will be ignored. |
| small blue | large blue | |
| large red | small red | If jostling causes a large red field to lose contact with its bonded large blue field, then both fields switch to their small states and their codons are no longer designated as being bonded. |
| large blue | small blue | |
| small green | large green | If a codon's red or blue fields are bonded, then its green or purple field switches to its large state. |
| small purple | large purple | |
| large green | small green | If neither of a codon's red or blue fields are bonded, then its green or purple field becomes small. |
| large purple | small purple | |
| small yellow | large yellow | If a double strand is ready to split, then the yellow fields of all of the codons in the double strand become large (this is described in more detail later). |
| large yellow | small yellow | If a yellow field has been large for 150 time units, then it returns to its small state. |

Table 3. The behaviour of the fields. Fields have no effect on each other unless their circles intersect. If the behaviour of a pair of fields is not listed in this table, it means that pair of fields has no interaction (they ignore each other). The designations "Field 1" and "Field 2" in this table are arbitrary, since the relationships between the fields are symmetrical.

| Field 1 | Field 2 | Interaction between fields |
| --- | --- | --- |
| small red | small blue | If a small blue field touches a small red field and the arms of their respective codons are aligned linearly to within ± π/256 radians, then both fields switch to their large states and their codons are designated as being bonded together. As long as the two fields continue to intersect, at any angle, they remain bonded, and any third field that intersects with the two fields will be ignored. |
| large red | large blue | When a large red field is designated as bonded with a large blue field, an attractive force pulls the tip of the red arm towards the tip of the blue arm and a straightening force twists the codons to align their arms linearly. |
| small purple | small green | When a purple field touches a green field and the arms of their respective codons are aligned linearly to within ± π/3 radians, they are designated as being bonded. As long as the two fields continue to intersect, at any angle, they remain bonded, and any third field that intersects with the two fields will be ignored. An attractive force pulls the tip of the purple arm towards the tip of the green arm and a straightening force twists the codons to align their arms linearly. When a small purple field bonds with a small green field, their bond is typically quickly ripped apart by brownian motion. The bonds between two large fields or one large field and one small field are more robust; they can withstand interference from brownian motion. |
| small purple | large green | |
| large purple | small green | |
| large purple | large green | A large purple field and a large green field cannot initiate a new bond; they can only maintain an existing bond. If they do not have an existing bond, carried over from before they became large, then they ignore each other. Otherwise, as long as the two fields continue to intersect, at any angle, they remain bonded, and any third field that intersects with the two fields will be ignored. An attractive force pulls the tip of the purple arm towards the tip of the green arm and a straightening force twists the codons to align their arms linearly. |
| large yellow | large yellow | When two large yellow fields intersect, a repulsive force pushes them apart. The repulsive force stops acting when the fields no longer intersect or when the fields switch to their small states. However, when the repulsive force stops acting, the codons will continue to move apart, until their momentum has been dissipated by the viscosity of the liquid. For yellow fields, unlike the other fields, there is nothing that corresponds to designated bonded pairs. Thus, if there are three or more intersecting large yellow fields, they will all repel each other. |

Table 4. Transition rules for *strand_location_state*. When *strand_location_state* is 2, the given codon must actually be at one end of a (possibly incomplete) double strand. During the replication process, if a strand has not yet fully replicated, there will be gaps in the strand, and the codons situated at the edges of these gaps will be stuck in state 1 until the gaps are filled, at which time they will switch to state 0.

| Current state | Next state | Transition rules for *strand_location_state* |
|---|---|---|
| 0 | 1 | If (I have exactly one red or blue neighbour) and (I have a purple or green neighbour), then I switch from state 0 to state 1. |
| 1 | 0 | If (I do not have exactly one red or blue neighbour) or (I do not have a green or purple neighbour), then I switch from state 1 to state 0. |
| 1 | 2 | If (I have exactly one red or blue neighbour) and (I have a green or purple neighbour) and (my green or purple neighbour is in state 1 or 2), then I switch from state 1 to state 2. |
| 2 | 0 | If (I do not have exactly one red or blue neighbour) or (I do not have a green or purple neighbour) or (my green or purple neighbour is in state 0), then I switch from state 2 to state 0. |

Table 5. Transition rules for *splitting_state*. A strand begins with all codons in the x state. When the strand is complete, one end of the strand (the end with no red neighbour) enters the y state, and the y state then spreads down the strand to the other end (the end with no blue neighbour). If the double strand is incomplete, the codons next to the gap will have their *strand_location_state* set to 1, which will block the spread of the y state. When the y state spreads all the way to the other end, in either of the two single strands, the double strand must be complete. Therefore, when the y state reaches the other end (the end with no blue neighbour), the end codon enters the z state, and the z state spreads back down to the first end (the end with no red neighbour).

| Current state | Next state | Transition rules for *splitting_state* |
| --- | --- | --- |
| x | y | If [(my *strand_location_state* is 2) and (my green or purple neighbour's *strand_location_state* is 2) and (I have no red neighbour)] or [(my *strand_location_state* is not 1) and (my green or purple neighbour's *strand_location_state* is not 1) and (my red neighbour's *splitting_state* is y)], then I switch from state x to state y. |
| y | z | If [(my *strand_location_state* is 2) and (my green or purple neighbour's *strand_location_state* is 2) and (I have no blue neighbour)] or [(my *strand_location_state* is not 1) and (my green or purple neighbour's *strand_location_state* is not 1) and (my blue neighbour's *splitting_state* is z)], then I switch from state y to state z. |
| z | x | If [(I have no red neighbour) and (I have been in state z for 150 time units)] or [my red neighbour is in state x], then I switch from state z to state x. |

Table 6. The parameters in JohnnyVon and the values that were used in the experiments.

| Parameter name | Parameter value | Description |
| --- | --- | --- |
| *timestep_duration* | 0.15 | See Section 3.7. |
| *linear_viscosity* | 1 - power(1 - 0.10, *timestep_duration*) | See Section 3.5.2. |
| *angular_viscosity* | 1 - power(1 - 0.05, *timestep_duration*) | See Section 3.5.2. |
| *linear_spring_damping* | 1 - power(1 - 0.90, *timestep_duration*) | See Section 3.5.6. |
| *angular_spring_damping* | 1 - power(1 - 0.99, *timestep_duration*) | See Section 3.5.6. |
| *iterations_after_split* | (integer) (150.0 / *timestep_duration*) | See Section 3.6. |
| *arm_length* | red arm: 7, blue arm: 7, green arm: 4, purple arm: 4, yellow arm: 1 | See Table 1. |
| *small_field_radius* | all fields: 0.01 | See Table 1. |
| *large_field_radius* | yellow field: 6, other fields: 4 | See Table 1. |
| *arm_force* | red arm: 1.8, blue arm: 1.8, green arm: 1.0, purple arm: 1.0, yellow arm: 1.0 | The strengths of the fields; see Sections 3.5.3 and 3.5.4. |
| *angle_tolerance* | red arm: ± π/256 radians, blue arm: ± π/256 radians, green arm: ± π/3 radians, purple arm: ± π/3 radians, yellow arm: no limit | See Tables 2 and 3. |
| *straightening_force* | red arm: 1.0, blue arm: 1.0, green arm: 0.5, purple arm: 0.5, yellow arm: none | See Section 3.5.5. |

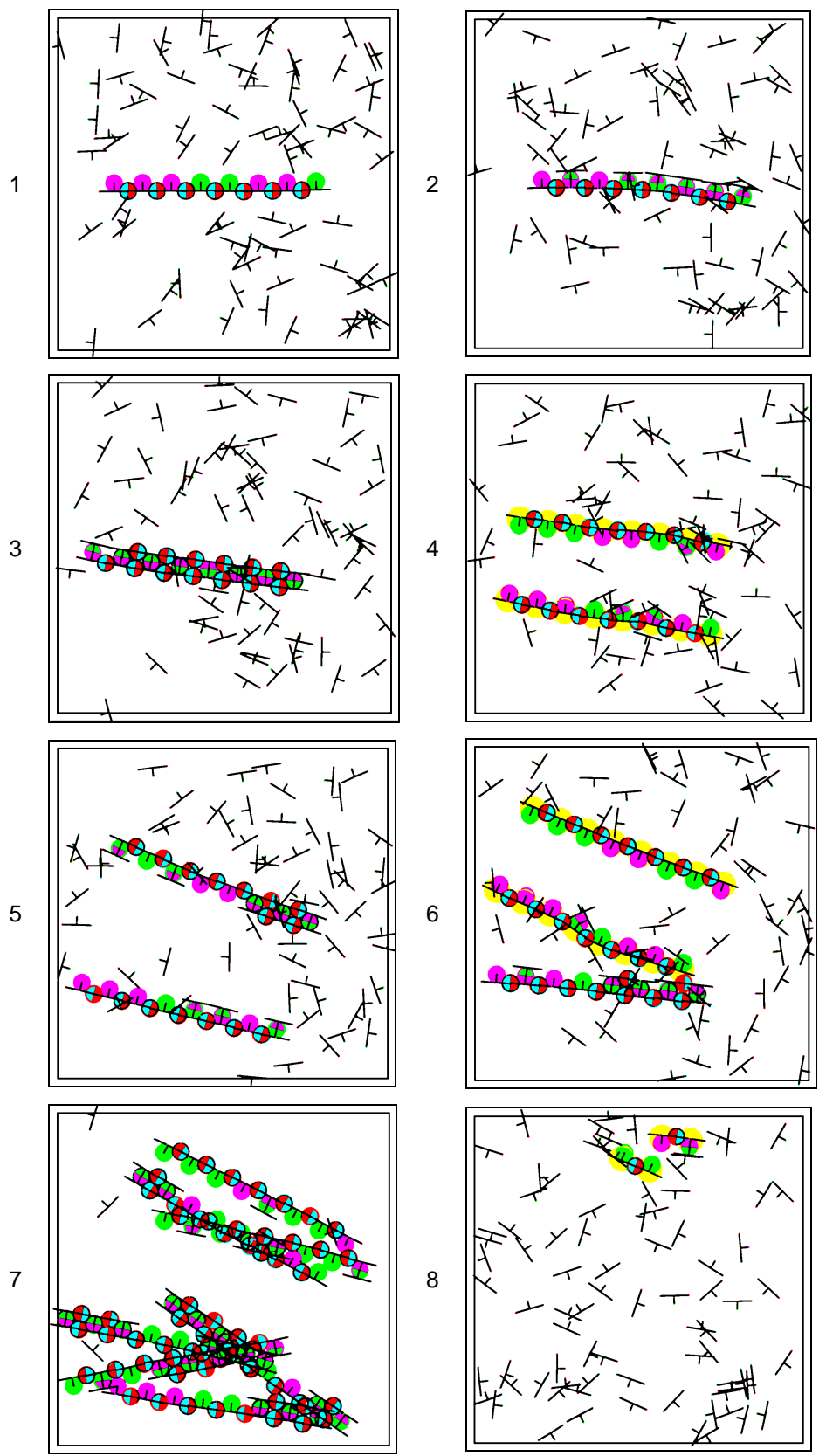

Figure 1. These images illustrate the experiments in Sections 4.1 and 4.2. See the text for details.